# Deep Neural Networks – A Brief History

Krzysztof J. Cios
Virginia Commonwealth University and IITiS Polish Academy of Sciences

**Introduction**

In this article we describe Deep Neural Networks (DNN), their history, and some related work.

DNN are one of the most efficient tools that belong to a broader area called deep learning. DNN process input information in a hierarchical way, where each subsequent level of processing extracts more abstract /global / invariant features. In other words, DNN (semi) automatically learn key features from data and then aggregate them for some purpose, such as recognizing objects in the images.

We shall illustrate how DNN work by the use of an example from the area of face recognition. There, the inputs are images from which at the first level (first hidden layer) of processing simple image characteristics such as edges are extracted. At the second and subsequent levels, more complex parts of an image are formed to finally, at the output layer, recognize human faces. This is in contrast to using a traditional approach where in the first step, known as preprocessing, an expert guides the process of extracting key features, and then they are used for recognizing faces. The common part of these two, very different, approaches is that at the output layer the labeled data are needed to perform supervised learning, i.e., assigning names/labels to faces.

Although DNN can in general work in all three basic learning modes, namely, supervised, unsupervised, and semi-supervised, so far the majority of successful DNN applications used the semi-supervised mode where (almost) unsupervised extracting of key features by the hidden layers was followed by a supervised learning at the output layer. In the fully supervised DNN mode the most frequently used algorithm is backpropagation with a ramp/rectifier activation function, which is very efficient in networks with many layers. The supervised approach, however, contradicts the very idea of deep learning as it is just a classical backpropagation learning with a sigmoid replaced by the ramp function, $f(x)=\max f(0,x)$. At the other end of the spectrum, in the area of fully unsupervised DNN, little progress has been reported so far.

DNN, as well as other types of neural networks, were inspired by the need to solve difficult for computers problems, such as image recognition but that are easily solvable by humans. Specifically, they were inspired by our, although still very vague, understanding of how human brain processes information. Depending on a goal of brain modeling we distinguish two approaches. If the goal is to model brain's neural circuits, the area called neuroinformatics or computational neuroscience, a key question validating the generated model is: How well does it fit the experimental biological data? In this approach, the neuron model frequently used is the spiking one with the appropriate learning rule. On the other hand, if the goal is to solve a practical problem, such as face recognition, then the validation question changes to: Is the model efficient? As in the latter case it is not important whether a simple or complex neuron model or any specific learning rule is used. This type of modeling is known as neuromorphic computing.

A digression about capacity of a human brain. It has about $10^{11}$ neurons and trillions of synaptic connections, which endows it with enormous storage capacity. If we define storage capacity as the ratio of the number of patterns that can be stored and retrieved, to the size of the network, then a network consisting of N neurons can retrieve correctly P stored patterns, according to this formula: $P < N/(4*\ln(N))$. Thus, a network with $10^4$ neurons can store only 271 patterns and with $10^{11}$ neurons it grows to $10^9$ patterns. The latter number of patterns is more than enough for a human to store and remember every single image, word, situation etc. encountered during a lifetime. In fact, the human brain has even bigger storage capacity because a group of neurons can store not just one but many different patterns, the phenomenon known as polysynchrony (Izhikevich, 2006). Fortunately, most people do not remember everything from the time they are born, although there are well documented cases of individuals who remembered everything from their past, day by day. By comparison, current artificial neural networks are incomparably smaller, with the largest using up to tens of thousands of neurons. One of the reasons for the size is that neural networks are designed to solve domain-specific problems, versus solving problems for many domains at the same time. For example, one network is designed to solve an image recognition problem while another a natural language processing problem, but there are no attempts to design a single network for solving problems from both domains.

Our focus here is on DNN, including those that use spiking neuron models and the corresponding learning rules. We start by defining key building blocks of all DNN. They are: a) a neuron model, which performs basic computations, b) a learning rule, which updates the weights/synapses between the neurons, and c) a network architecture, which specifies how the neurons are topologically arranged and interconnected.

**Neuron Models**

A wide spectrum of neuron models from very simple to spiking ones is described next. Notice that increasing biological detail of an artificial neuron model also increases its computational complexity.

The first simple model of a neuron, called the *threshold neuron*, was developed by McCulloch and Pitts (1943). It calculates a dot product between the input vector and the weight vector of a neuron, and if it is higher than its transfer function (like step function) it fires/generates an output of 1 (otherwise 0).

The first *spiking neuron* model was developed by Hodgkin and Huxley (1952), for which they later received a Nobel Prize. They modeled squid's giant neuron and they treated each component of the neuron, including its membrane, as electrical component. The model is described by:

$$C \frac{dV}{dt} = I_e - \overbrace{\overline{g}_K n^4 (V - E_K)}^{I_K} - \overbrace{\overline{g}_{Na} m^3 h (V - E_{Na})}^{I_{Na}} - \overbrace{g_L (V - E_L)}^{I_L}$$

where:
$I_e$ = stimulus / injected current
$V$ = voltage/membrane potential
$L$ = leakage current
$K$ = potassium and $Na$ = sodium channels
$g$ = conductances, e.g., $g_{Na}$=120 mS/cm$^2$; $g_K$=36 mS/cm$^2$; $g_L$=0.3 mS/cm$^2$
$E$ = reversal potentials, e.g., $E_{Na}$=115mV, $E_K$=-12 mV, $E_L$ = 10.6 mV
$n, m, h$ = channel gating/activation variables: $n=n(t), m=m(t), h=h(t)$

To better understand it let us look at its equivalent electric circuit, shown in Figure 1. The membrane is modeled as capacitor, $C_m$, while potassium and sodium ion channels as conductances (g = 1/R; R being resistance). V is the neuron's membrane potential, i.e., difference between its intracellular (inside of the neuron) and extracellular potentials. According to Kirchhoff's law the sum of the currents is zero so the current through the membrane, C dV/dt, can be written in a shorter form as:

C dV/dt = $I_e$ - $I_K$ - $I_{Na}$ - $I_L$

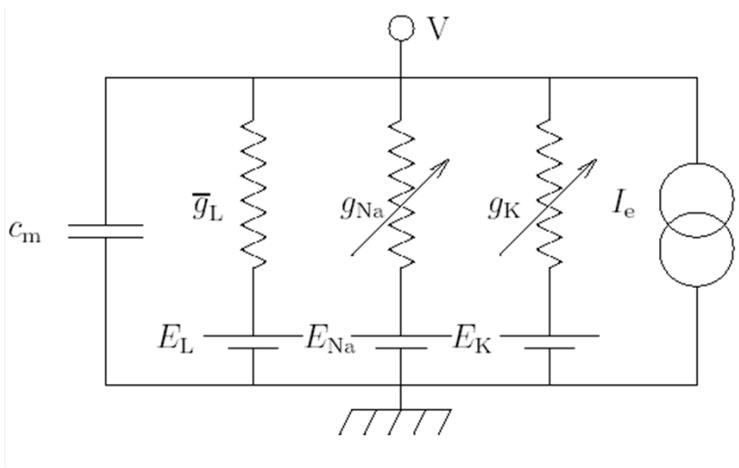

Figure 1. Hodgkin and Huxley's model circuit representation.

Figure 2 Illustrates generation of an action potential /spike by the flows of sodium and potassium ions, represented here as conductances.

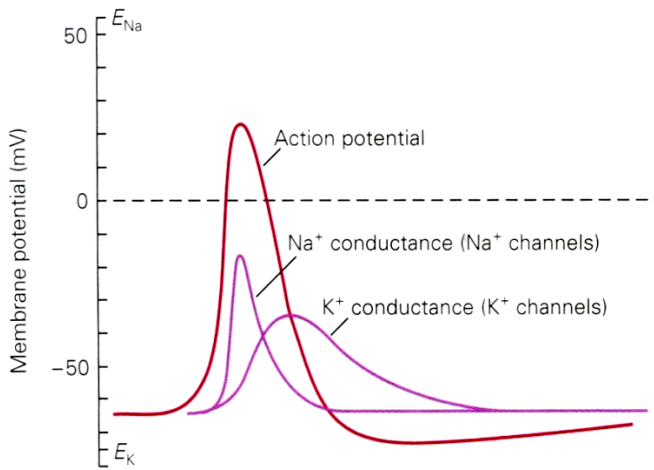

Figure 2. Generation of an action potential by sodium and potassium ions flows.

McGregor (1987) defined a simpler than Hodgkin-Huxley spiking neuron model, one that belongs to a group of *integrate-and-fire* models.

It is described by these equations:

$$S = \begin{cases} 1 & E \geq T_h \\ 0 & E < T_h \end{cases}$$

$$\frac{dE}{dt} = \frac{-E + G_K \cdot (E_K - E) + G_e \cdot (E_e - E) + G_i \cdot (E_i - E) + SCN}{T_{mem}}$$

$$\frac{dG_K}{dt} = \frac{-G_K + B \cdot S}{T_{GK}}$$

$$\frac{dT_h}{dt} = \frac{-(T_h - T_{h0}) + c \cdot E}{T_{Th}}$$

where:
- V – membrane potential
- $V_r$ – membrane resting potential
- $V_K$ – potassium resting potential
- $V_i$ – inhibitory resting potential
- $V_e$ – excitatory resting potential

Transmembrane potentials: $E = V - V_r$;  $E_K = V_K - V_r$;  $E_i = V_i - V_r$;  $E_e = V_e - V_r$

Transmembrane conductances: $G_K = g_K / G$;  $G_i = g_{si} / G$;  $G_e = g_{se} / G$

- G – membrane resting conductance
- $g_K$ – potassium resting conductance
- $g_{si}$ – inhibitory resting conductance
- $g_{se}$ – excitatory resting conductance
- $T_{GK}$ – decay of GK time constant
- $T_h$ – threshold value
- $T_{h0}$ – resting value of threshold
- $T_{th}$ – decay of threshold constant
- $T_{mem}$ – membrane time constant
- $T_{mem} = C/G$

Current through membrane:
SCN = SC/G
SC – current injected to cell (corresponds to $I_e$ in the HH model)
c  – rise of threshold $c \in [0,1]$
C  – membrane capacitance
B  – postfiring potassium increment

Its corresponding electric circuit, shown in Figure 3, is similar to Hodgkin and Huxley's. It models the potassium channel, refractory properties, adaptation to stimuli, and mimics excitatory and inhibitory post synaptic potentials (EPSP and IPSP, respectively) of a neuron. The PSPs are illustrated in Figure 4. We will refer back to these potentials when we later describe learning rules.

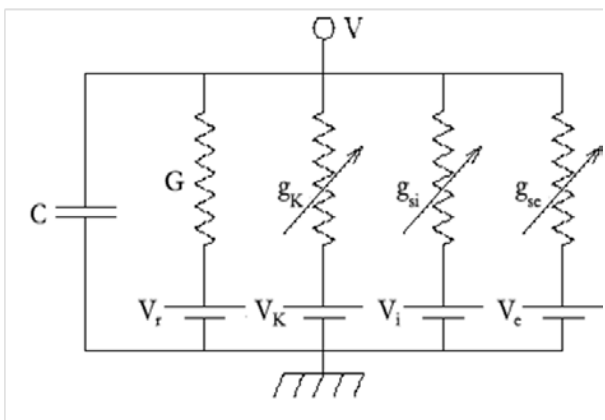

Figure 3. McGregor's model circuit representation.

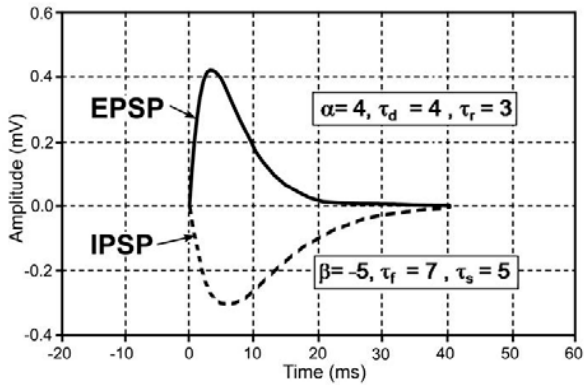

Figure 4. Example excitatory and inhibitory post synaptic potentials.

The working of the McGregor's model is illustrated in Figure 5, using a "network" of only three neurons: two pre-synaptic (one excitatory and one inhibitory) that feed into one post-synaptic neuron (Sala and Cios, 1999). We can see the spikes that are generated by both types of pre-synaptic neurons, shown in the bottom panel of Figure 5. The positive excitatory ($G_e$) and negative inhibitory ($G_i$) inputs feed into a post-synaptic neuron that integrates them and when the sum rises above its threshold (Th) the post-synaptic neuron fires a spike; four such spikes are generated by the post-synaptic neuron, which is shown in the top panel (E)) of Figure 5. Notice that the threshold of the neuron (Th) changes over time.

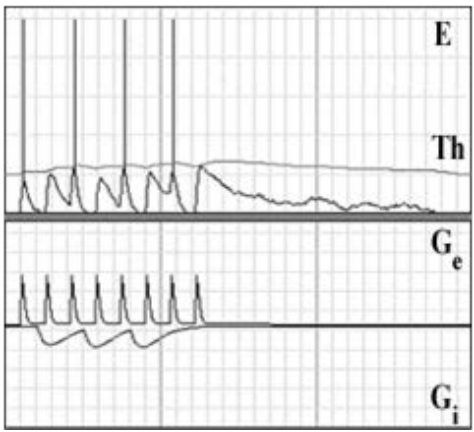

Figure 5. The input pre-synaptic signals ($G_e$ and $G_i$) make the post-synaptic neuron (E) to fire four spikes.

The simplest spiking neuron model was developed by Izhikevich (2003). It does not model any of the biological neuron functions except that it accurately mimics several types /shapes of the postsynaptic potentials generated by human brain neurons. It is described by:

$$\begin{cases} v' = 0.04v^2 + 5v + 140 - u + I \\ \quad u' = a(bv - u), \end{cases}$$

where
if v > 30, then v = c, u = u + d,
- v  is membrane potential
- u  is membrane recovery variable
- I   is input current

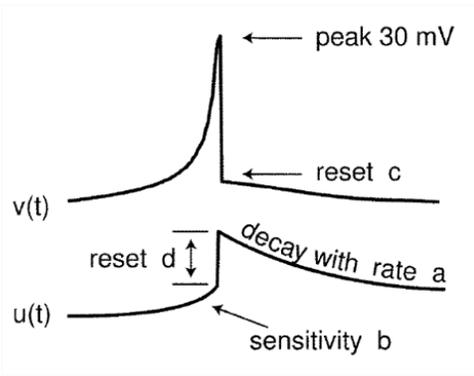

It models over a dozen different post-synaptic firing patterns, four of which are illustrated in Figure 6; they correspond to these parameter settings in the model:

| Parameter | Excitatory (RS, IB) | Inhibitory (FS,LTS) |
|---|---|---|
| a | 0.02 | $0.02 + 0.08\gamma$ |
| b | 0.2 | $0.25 - 0.05\gamma$ |
| c | $-65 + 15\gamma$ | -65 |
| d | $8 - 6\gamma$ | 2 |
| $\gamma$ is a uniform random variable between 0 and 1 | | |

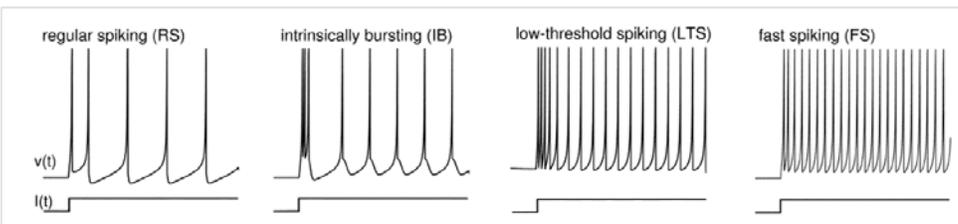

Figure 6. Two types of excitatory (the first two) and two types of inhibitory neurons firing patterns (taken from http://www.izhikevich.org/publications/spikes.htm).

Izhikevich model became very popular because its simplicity allows for building networks consisting of thousands of such neurons. While using it, however, we found that increasing the strength of the stimulus caused it to fire with higher and higher frequency (no upper bound). This is not biologically plausible as neurons cannot fire during the absolute refractory period, needed for restoration of their membrane potentials, no matter the strength of the input. We thus corrected the condition for the neuron firing (Strack, Jacobs and Cios, 2013) by changing it
from
    *if v > 30, then v = c, u = u + d*
to
    *if v > 30:*
    *if*
*dt> dtmin: v = c, u = u + d,* **spike**
**else**
*v = 30,* **no spike**

That is, we added additional check (if dt>dtmin) to account for refractory property of neurons. Figure 7 A) below illustrates firings of the four types of original Izhikevich neurons, while Figure 7 B) shows firings of neurons after our modification. The modified model was used for modeling multi-column multi-layer model of neocortex, which was not possible to do using the original Izhikevich model (Strack, Jacobs and Cios, 2014).

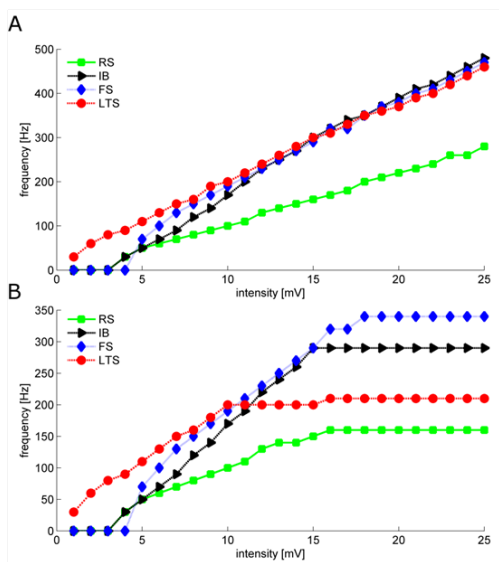

Figure 7. A) unbounded firing of the original Izhikevich model neurons; B) firing of the neurons after Strack at el. (2014) modification accounting for absolute refractory periods.

**Learning Rules**

Let us start by noticing that almost all learning rules are based to some degree on Konorski's observation:
*IF a presynaptic neuron "j" repeatedly fires a postsynaptic neuron "i " within a short time*
*THEN the synaptic strength between the two is increased, otherwise it is decreased.*

The credit for the above observation most often is given to Hebb (Hebb, 1949) although Konorski published it a year earlier (Konorski, 1948). The practical learning rules, i.e., equations corresponding to the above observation were specified much later by computational scientists (Song et al., 2000; Swiercz et al., 2006).

Similar case, of not giving credit to the original inventor, involves a popular *backpropagation learning rule* that was first specified by statisticians Robbins and Monroe (1951): they called it a stochastic approximation method. However, the credit for the rule in neural networks literature was given to Rumelhart, Hinton and Williams (1986) before it was found that Werbos (1974) specified the rule, also for neural networks, a dozen years before them.

The simplest learning rule, called *Perceptron*, for one-layer feed-forward neural networks, was defined by Rosenblatt (1957). Backpropagation rule is in fact the Perceptron's rule extension to many-layer networks. Extending it to such networks, however, became possible only after the step threshold function used in the Perceptron was replaced with a differentiable sigmoid function. This seemingly small change led to an explosion in neural networks research that stagnated for almost 20 years after Minsky and Papert (1969) stated that neural networks were useless for solving complex problems.

Kohonen (1982) specified *winner-takes-all learning* rule. This rule more closely than Perceptron or backpropagation mimics the learning processes taking place in biological neural circuits. It states that only the neuron whose weight vector (synapse) is the closest to the input's vector is the winner and as such increases its weight to get it even closer to the input pattern vector. Often, a number of neurons in close neighborhood of the winning neuron also adjust their weights.

The first rule for networks of spiking neurons, called *Spike Time-Dependent Plasticity* (STDP), was specified by Song, Miller and Abbot (2000). Swiercz, Cios, Staley et al. (2006) specified another rule for spiking neurons called *Synaptic Activity Plasticity Rule* (SAPR). The two rules are compared in Figure 8. Konorski's observation is translated, in both rules, into the following recipe:
*The adjustment of the strength of synaptic connections between pre-synaptic neuron "j" and post-synaptic neuron "i" takes place every time the postsynaptic neuron ''i" fires, according to the function specified either by STDP or SAPR. If Δt*

*is positive that means the pre-synaptic neuron fired **before** the post-synaptic neuron and the strength between the two is increased.  If Δt is negative it means that the pre-synaptic neuron fired **after** the post-synaptic neuron fired and the strength between the two is decreased.*

The difference between the two rules is that SAPR uses a function that is continuous and differentiable (important in several applications); it is also dynamic because it uses actual post-synaptic potential functions to modify the connection strengths between the neurons.  In other words, the adjustments depend on the shape of SAPR, which in turn depends on the shape of the chosen postsynaptic functions in a given neural circuit.  The left part of the SAPR function in Figure 8 (to the left of the y axis) is the chosen inhibitory PSP while the right part is the chosen excitatory PSP; see again the two function shapes in Figure 4.  In contrast, the STDP rule uses a static function meaning that the adjustments are always the same; they do not depend on the shape of inhibitory/excitatory PSPs for a given Δt.

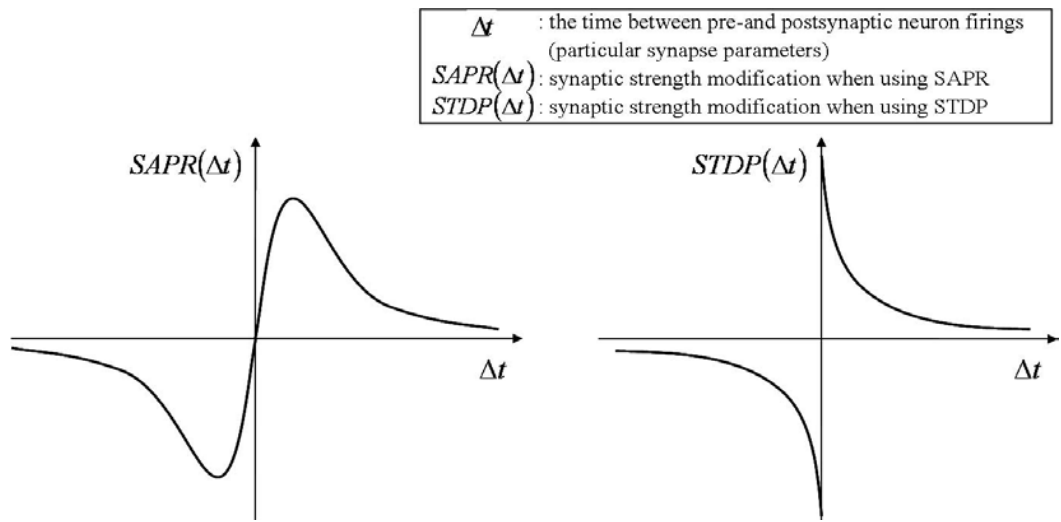

Figure 8. Comparison of the SAPR and STDP: the latter is fixed while the former changes depending on the shape of excitatory and inhibitory post synaptic functions of neurons.

**Network Architecture**

As stated above, DNN use a hierarchical architecture, vaguely mimicking the brain's hierarchical way of performing cognitive tasks. This architecture is one of the key distinguishing factors between several types of neural networks and DNN. It follows that neural networks with just one hidden layer, such as SVM, RBF, or Kohonen's self-organizing feature map, are not DNN.  As a digression, a popular decision tree algorithm does not perform deep learning either, in spite of its hierarchical architecture, since it uses original features and not a hierarchy of transformed features.

Hierarchical processing of information in the brain was first discovered by neurophysiologists Hubel and Wiesel (1962) who studied the cat's visual system; for this work they were awarded a Nobel Prize. Not only they observed the brain's hierarchical way of processing information but also that at each level of processing the brain extracts more general features performed by *complex* cells, that aggregate the features extracted at the previous level to, at the end of this process, recognize some objects in the input image.  At the first level, the brain focuses on recognizing specific simple patterns in the input images, such as vertical or horizontal elements present in input images, which are extracted by *simple* cells. Hubel and Wiesel were thus originators of the key ideas leading to development of DNN. It is easy to notice, see Figures 9 and 10, that the DNN of today use very similar architectures.

We explain Hubel and Wiesel's work in some detail using Figure 9.  A very simple model of the cat's visual processing system can be implemented using neuron model of McCulloh and Pitts, which outputs/fires a 1 when the sum of its inputs is above its threshold, and outputs a 0 otherwise.  By changing its threshold value, the neuron can perform logical operations of conjunction and disjunction. A conjunction is achieved as follows: if the threshold is relatively high, say, 3, then inputs from 3 presynaptic neurons (of 1 each) are required to fire it. Such neurons can recognize different line orientations in the images, such as vertical, horizontal, or diagonal. The neuron can also perform a disjunction if its threshold is relatively low, say, 1; then the input (of 1) from any of the three presynaptic neurons fires it. This is illustrated in Figure 9, where in the first column we see image of digit 2. The four neurons, the simple cells, in the first (hidden) layer perform conjunctions to recognize three-element line patterns, while the four neurons, the complex cells, in the second layer perform disjunctions that aggregate the simple patterns into more complex ones until, at the output layer, digit 2 is recognized. In the parlance of DNN the conjunction is called convolution, the disjunction a spatial pooling, the simple cell a feature extractor/detector, and the complex cell a feature aggregator/analyzer. The difference between the just described very simple scheme and DNN is that feature extraction in DNN happens without (almost) human intervention (we describe later how DNN do it).

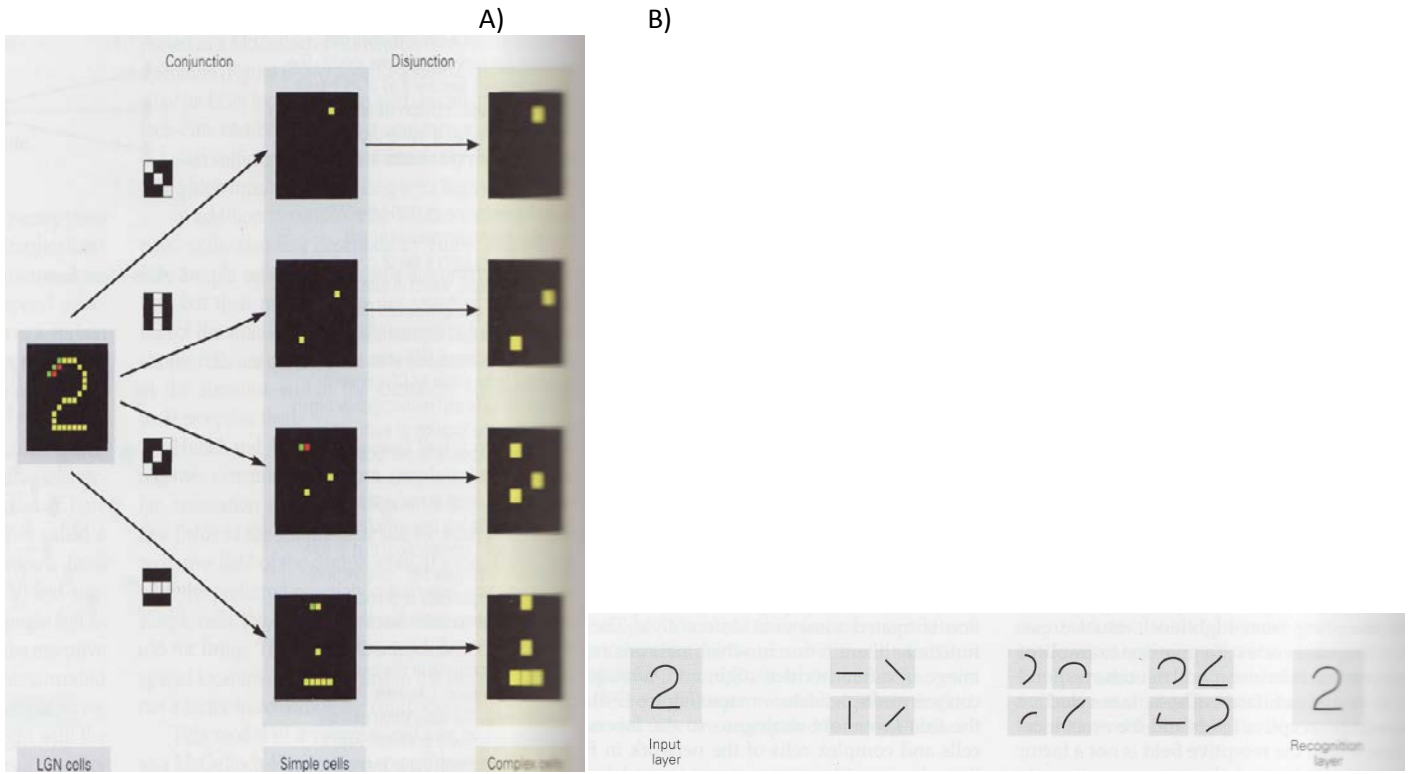

Figure 9. A) illustration of how the simple and complex cells extract specific features from input images; B) implementation of how the features are extracted and aggregated (using three hidden layers) in Neocognitron to recognize digit 2 (both pictures are taken from Kandel et al. Principles of Neural Science, 5$^{th}$ edition, 2013).

The first researcher to design a direct precursor of DNN, using Hubel and Wiesel's discoveries, was Fukushima (1980) who called his network Neocognitron. Figure 10 A) illustrates how key features of an image of letter A are first picked up by simple cells (S) and then aggregated by complex cells (C), in order to recognize letter A at the output. S-layer of simple cells extracts features from the previous stage in the hierarchy, while the C-layer of complex cells ensures tolerance for shifts of features extracted by the S-layer.

A

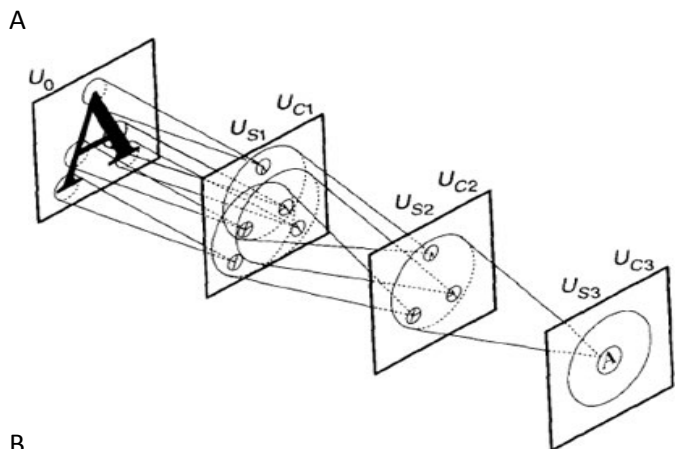

B

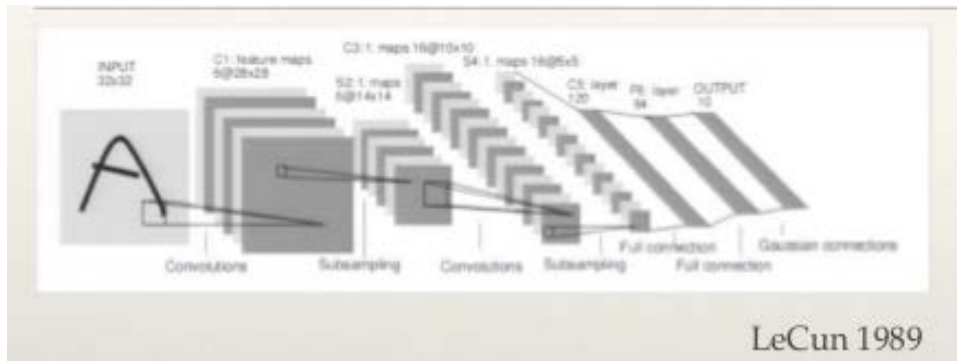

Figure 10. A) Fukushima's Neocognitron architecture, and B) LeCun's convolutional neural network architecture.

DNN became popular and the term "deep learning" was coined and widely accepted around 2010 due to the development of efficient learning algorithms and hardware speed-ups such as the use of GPUs. In particular, LeCun (1998; 2010; 2015), Hinton (2006a; 2006b) and Krizhevski (2012) made significant impact on the field. Comparison of architecture of Neocognitron shown in Figure 10 A) with the DNN architecture of LeCun's convolutional network shown in Figure 10 B) shows their great similarity.

As aforementioned, the first few layers of DNN perform feature extraction using unsupervised learning, and only the top layer weights (i.e., those between the last hidden and output layer) are trained in a supervised mode. In DNN the most often used approach to perform feature extraction between the input and hidden layer(s) is to use an idea of autoencoder. Other method often used to perform unsupervised learning (always a form of clustering) is a Boltzman machine.

We now explain how an autoencoder works using a feed-forward neural network with backpropagation learning in a vertical composition, meaning that the same operation that is performed by the first hidden layer on the original input, is also performed by the second layer on the output of the first hidden layer, etc. Let us also assume that our input is an image of size nxn and that the number of neurons in the first hidden layer is p, with p smaller than $n^2$ (this condition is not required but using it makes it easier to understand and explain). The task of the autoencoder is to learn outputs of the first hidden layer in such a way that after learning we can reconstruct (using outputs of the hidden layer) the inputs with a very small distortion. In other words, the autoencoder learns a compressed (lower dimensional) version of the inputs. In that respect, the autoencoder is similar to PCA and performs clustering.

Loosely speaking, the outputs of the first hidden layer neurons are trained to recognize some specific features, as linear combinations of the original image features, such as edges, in different positions and orientations. This is what is meant by saying that new features are automatically learned/extracted by deep neural networks. The same process is repeated at the output of the second hidden layer, which takes as input the output of the first hidden layer. The outcome of doing it is that the previously extracted features, say, edges, are aggregated into more complex features, say silhouettes of objects. Supervised learning is only then used to train the weights between the last hidden and the output layer in order to assign labels to the input images.

Instead of describing Neocognitron or convolutional neural network of LeCun for which many excellent online resources exist, we describe below a network called IRNN (Image Recognition Neural Network), which was inspired by the works of Hubel and Wiesel and Fukushima (Cios and Shin, 1995). In the IRNN the hidden layers perform explicit clustering operations of the (sub) images for the purpose of extracting key features at each level of hierarchical processing. IRNN consists of an input layer, an output layer, and one or more hidden layers, as shown in Figure 11. The Sensory layer extracts local features from the images. The role of the hidden layer(s) is to aggregate local features to generate higher level semi-global features. The output layer, in a supervised mode, associates the semi-global features with the known labels. Notice that IRNN operates like a semi-supervised convolutional DNN. It uses windowing, which is based on a biological observation that a neuron connected to the sensory system receives inputs from only a portion of the sensory neurons.

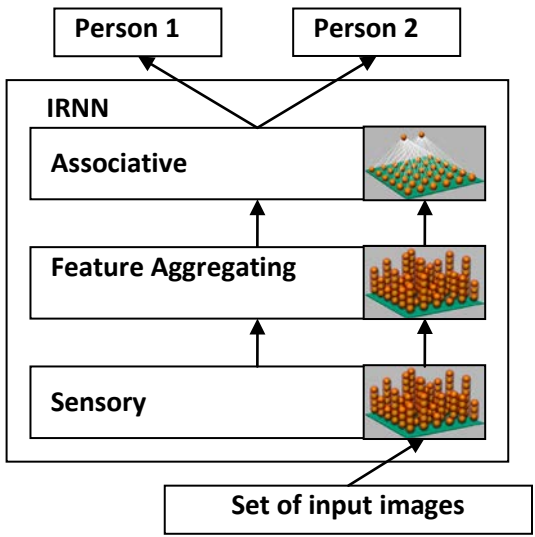

Figure 11. IRNN's architecture: unsupervised part consists of the sensory and feature aggregating layers while the associative part is supervised.

Figure 11 shows stacks of neurons represented by small balls. How they are generated and what they represent is explained in Figure 12. We see there three (hashed) subimages/windows of the three input images, which are clustered using a novel image similarity measure (Cios and Shin, 1995). If the first two subimages are similar (as shown) they are clustered together in neuron $n_1$. Since subimage 3 was found quite different from the subimages 1 and 2, it creates its own cluster, so the second neuron, $n_2$, is generated. The weights $w_1$ and $w_2$ are initially set to the first subimage pixel values vector but are later updated to represent cluster center (thus representing an "average" subimage). At the end of scanning of entire images the result might be as the one shown in Figure 13. Notice, that at the center more neurons (clusters) were created to represent image details, such as nose, eyes and mouth, while at the periphery where background was about the same in all images only single neurons/clusters were needed.

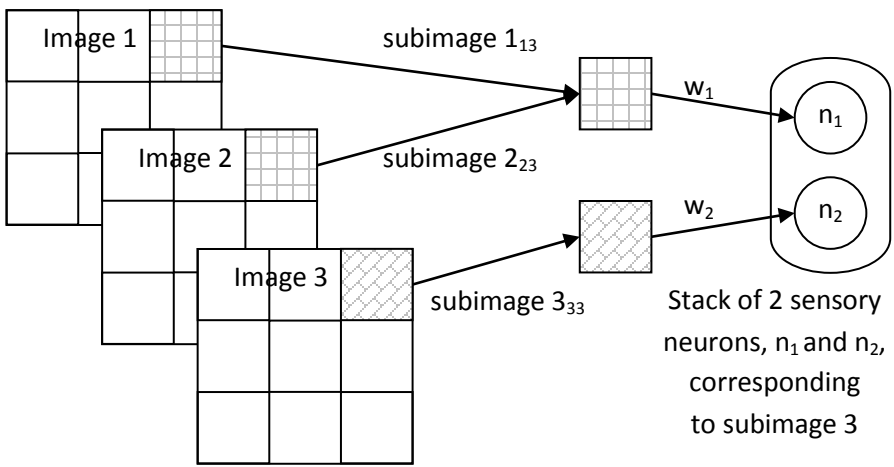

Figure 12. Explanation of clustering of subimages into a number of clusters/neurons.

| 1 | 1 | 1 | 1 | 1 | 1 | 1 | 1 |
|---|---|---|---|---|---|---|---|
| 1 | 1 | 2 | 2 | 3 | 1 | 1 | 1 |
| 1 | 2 | 2 | 3 | 4 | 2 | 2 | 1 |
| 1 | 2 | 4 | 5 | 3 | 2 | 1 | 1 |
| 1 | 1 | 3 | 6 | 5 | 3 | 1 | 1 |
| 1 | 2 | 4 | 4 | 5 | 3 | 1 | 1 |
| 1 | 1 | 2 | 2 | 2 | 1 | 1 | 1 |
| 1 | 1 | 1 | 1 | 1 | 1 | 1 | 1 |

Figure 13. A hypothetical result of clustering of a set of registered face images.

The same process is repeated on the outputs of the sensory layer to aggregate the local features into more complex semi-global features. Clustering of subimages at each level is performed with all other layers disabled. Finally, the associative layer associates the images with recognition codes – like Person 1 or Person 2 - using winner-takes-all learning rule. Note that the number of clusters/neurons in the IRNN is not predetermined by the user: it depends only on a similarity between the subimages. Characteristic feature of the IRNN is that if new data become available the already trained network can be used in two ways. If the new data points are labeled then additional training continues with the new data. However, if new data are not labeled only the output layer needs to be trained.

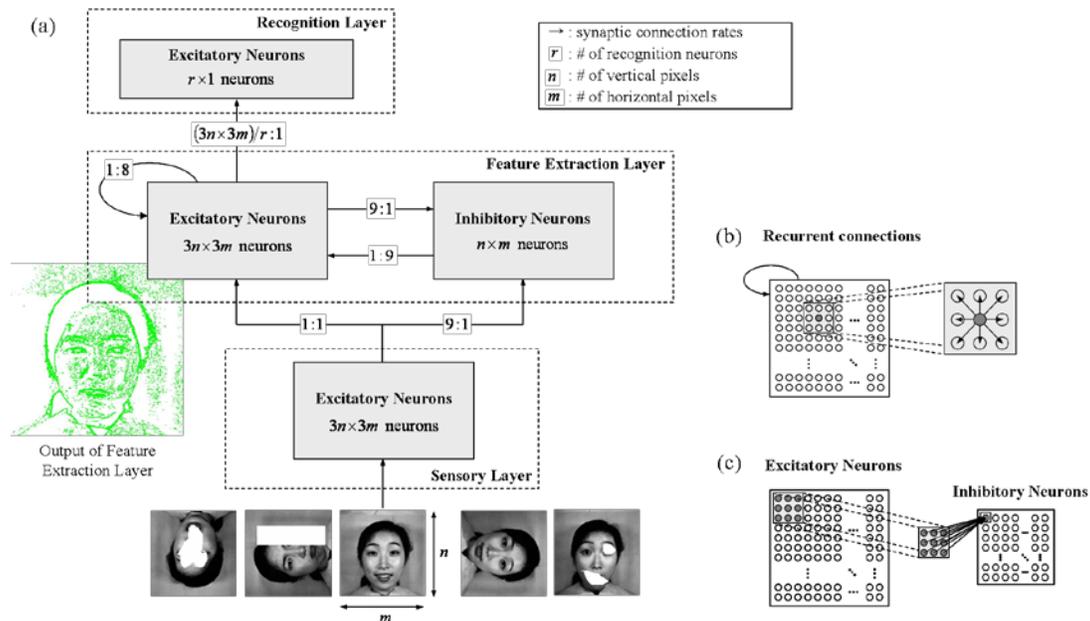

Figure 14. Architecture of the network of spiking neurons; (a) High-level block diagram. (b) Recurrent synaptic connections between the excitatory neurons in the feature extraction layer. (c) Synaptic connections between the excitatory neurons in the sensory/feature extraction layer and the inhibitory neurons in the feature extraction layer (taken from Shin et al. 2010 paper).

The networks described so far, including convolutional DNN, used only simple, non spiking, neuron models as their basic processing units. But is it possible to perform deep learning using networks of spiking neurons? Shin et al. (2010) used such a network for face recognition, without any preprocessing of the images. The network self-organizes at each level of its hierarchical processing. Even at the output layer spiking neurons are used for labeling faces, in contrast to more popular use of supervised methods such as backpropagation; similar approach was later used in Cao, et.al. (2014). Specifically, the spiking neuron model used was McGregor's with SAPR and STDP learning rules for self-organization of the neurons; self-organization in essence being a clustering operation. Figure 14 shows architecture of this a network. The sensory layer serves as relay of the input image but increases the input dimension, from an image of size nxm to an image of size 3nx3m. The only hidden layer, the Feature Extraction layer, is composed of excitatory and inhibitory neurons, while keeping their ratio close to the one observed in human brain. Notice that no supervised training is performed by this layer. Instead, multiple submitting of the input images is required until there is a negligible change in the result of self-organization. It was shown that using SAPR rule gave better results in recognizing face images than using STDP. The Recognition layer also uses spiking neurons: the neuron that spikes the most for a known input face image "recognizes" the person. The network performed particularly well on rotated and partially occluded images. In short, the network uses raw images for input, extracts key features without any training between the sensory and feature extracting layers. This is in contrast to using an autoencoder that can be seen as a supervised training method where the label is the compressed pattern of the input pattern.

**Problems with DNN learning**

Popular literature paints the advent of DNN as the panacea for solving difficult problems, such as image recognition, hand written character recognition, etc. Moreover, that it is done with high confidence/accuracy and without the need for human participation. Unfortunately, history of science tells us that new technologies are often accompanied by a high dose of hype, and DNN are no exception to this. As described below, two groups of researchers have shown spectacular failing of DNN on image recognitions tasks that are trivial for humans.

In one experiment, the researchers used a trained DNN and ran it on slightly modified images, called adversarial examples. The network has seen the original images (before modification) in training. The modification was such that that there was no perceptible to the human eye difference between the original and adversarial image; the latter had only slightly different statistical properties. For example, there was no way to tell the difference between the image of a dog and its slightly modified image. However, when the latter was input to the AlexNet (open source implementation of a convolutional neural network), which was trained on the original image of the dog, it failed to recognize it (Szeged et al., 2014).

In another work, the researchers took just the opposite approach. Namely, they modified (using genetic algorithms) the image used in training in such a way that it had no resemblance whatsoever to the original image. For example, an image looking like a TV static noise was not only recognized by LeNet (part of Caffe software package) say, as peacock, but also was very certain (accuracy of 99.6%) about its recognition decision(Nguyen, Yosinski and Clune, 2014).

**Conclusions**

The described above DNN shortcomings do not outweigh their many advantages. However, lots of research is needed to answer the question of why they failed in those experiments. I think it is increasingly more important for the computational researchers to team up with neuroscientists to come up with better algorithms for image recognition so that the algorithms cannot be so easily fooled (Lim et al., 2011). That may require, in the first place, more work by the neuroscientists to better understand processes used by the brain in recognition tasks.

The easy fooling of DNN in some recognition tasks, which are easily recognized by humans, poses a very serious cybersecurity risk. Modern society heavily relies on machine learning techniques, like DNN, for performing many everyday tasks such as medical diagnosis, self-driving cars, investing financial assets, and even in a legal system. Since the researchers have shown that it is relatively easy to come up with adversarial examples, the automated systems we so much now depend on can produce possibly disastrous results. It is thus increasingly important for researchers to add safety features to new deep learning algorithms they are developing, something that software engineers have been doing for a long time to assure safety of their code. To start with, researchers should routinely use in training adversarial examples, in addition to original ones, to make their systems more secure.